# Federated Learning Framework for Privacy-Preserving Kidney Stone Detection

**NAJIYYA YOUNAS[1], OMAR ABDULKADER[2], YASER ALI SHAH[1] , MUHAMMAD JAWAD IKRAM[2], JEBRAN KHAN[3,*], AMAAD KHALIL[4]**

[1]Department of Computer Science, COMSATS University Islamabad, Attock Campus, Attock 43600, Pakistan

[2]Faculty of Computer Studies, Arab Open University, Riyadh, Saudi Arabia

[3]Department of Artificial Intelligence, Kyungdong University Global, Toseong-myeon 24764, South Korea.

[4]Department of Computer Systems Engineering , University of Engineering & Teschnology,Peshawar, 25000, Pakistan.

Corresponding author: Jebran Khan (e-mail: jebran@kduniv.ac.kr).

**ABSTRACT** Recent innovations in deep learning have significantly enhanced the diagnosis of medical images, although they are based on the use of centralized data storage that pose severe threats to patient privacy and medical data security. To address this issue, this research proposes a Federated Learning (FL) model that is coupled with an optimized YOLOv8 network to detect the kidney stones on a computed tomography (CT) image and at the same time, protect privacy of the patients. The suggested system can help various medical organizations to jointly train a common model without exchanging the information about the patients. This is to ensure that data protection laws like GDPR and HIPAA are adhered to. The residual feature fusion and DropBlock regularization among other architectural improvements are also included in YOLOv8 to enhance detection robustness and minimize overfitting. Experimental analysis carried out on a distributed CT dataset demonstrated that the federated YOLOv8 model has a mAP at 50 of 0.733 and is able to keep the data confidential. Moreover, its lean design facilitates fast edge deployment and real-time inference across a clinical setting. Altogether, these findings indicate that Federated Learning is a safe and efficient solution to AI-assisted diagnosis in contemporary healthcare when combined with the use of sophisticated object detection models.

## I. INTRODUCTION

One of the most prevalent urological conditions that has impacted approximately 12 percent of the global population is kidney stone disease [1]. Cases of the affected have tremendously been on the rise in the past twenty years. This increase in cases has several causes which are: change in food habits, physical inactivity and genetic predisposition. In case the disease remains untreated, the disease may cause severe complications, such as hydronephrosis and frequent urinary tract infection. In worst cases, it may lead to permanent kidney damage as well. Early diagnosis is important because it will enable the physicians to apply less invasive measures like extracorporeal shock wave lithotripsy (ESWL) or ureteroscopy before any more complicated measure is required [2].

Medical imaging is essential in providing the accurate diagnosis of kidney stones. CT scans have high sensitivity and specificity, thus, they are known as the gold standard, however, patients are subjected to ionizing radiations, which can be dangerous over time as a result of recurrent screening [3]. Ultrasound imaging on the contrary, although very affordable, safe, and widely available, depends very much on operator expertise and machine parameters to achieve diagnostic performance. Stones less than 3 mm of diameter are particularly difficult to locate in the presence of noisy ultrasound in the background. Hence, automated detection systems play a very significant role in improving the reliability of the diagnosis and reducing the variability in observers [4].

Emerging technologies in deep learning have transformed the sphere of medical image analysis radically. Other models such as Convolutional neural networks (CNNs) and U-Net [5], [6] have demonstrated strong outcomes in the categorization and separation of medical images. Such models have successfully been applied to different fields e.g. pathology, radiology and dermatology [7], [8].

A range of ML and DL techniques have been researched in the field of kidney stones detection. Both approaches have their own benefits and difficulties. Neural networks, support vectors machines (SVM), and random forests (RF) are some of the most popular machine learning algorithms in this context. Studies have demonstrated that random forests and other techniques of ensemble learning have high classification accuracy in situations where they are used to provide high classification of multifaceted medical data which may have imbalanced or missing values. Likewise, convolutional neural networks (CNNs) are DL models that have demonstrated useful operation in image-based diagnostics, e.g. in automated kidney stones detection in CT scans and ultrasound images. It is worth noting that transfer learning on CNNs has been effective in enhancing the diagnostic performance and at the same time minimizing the requirement of large labeled datasets.

In spite of these developments, some obstacles remain in the use of ML and DL methods in normal clinical practice to detect kidney stones. The use of large, high-quality datasets that are contributed by many organizations is one of the significant barriers. HIPAA and GDPR, among other legal privacy-related regulations, frequently do not permit pooling of patient imaging data, and it is therefore infeasible to centrally train healthcare [9].

Federated Learning (FL) has become a potential solution to this problem [10]. In FL, every organization involved in the process trains a local model using its own private data and only transmits the weight changes to a central aggregator. Secure aggregation protocols will make sure that the server will not be able to reassemble sensitive information using these updates [11]. This would facilitate joint training and at the same time maintain patient confidentiality. Patient privacy is a very important issue in medical AI. FL can find particular application in privacy-relevant areas such as healthcare where strict privacy regulations such as HIPAA and GDPR are very limiting to the sharing of medical imaging and diagnostic data which is central to the creation of machine learning applications [12]. During the medical image analysis, FL allows a hospital and a research center to co-create robust and generalized models with the purpose of diagnosing and treating diseases without revealing the patient-sensitive information. FL simplifies the development of large-scale models that may utilize diverse datasets of diverse sources by being able to store the data locally. This enhances the accuracy and applicability of machine learning algorithms to various groups of patients.

Meanwhile, the object detection algorithms have developed rapidly, with the family of You Only Look Once (YOLO) being the most popular ones [13]. YOLO models can strike a balance between accuracy and real-time inference speeds and thus are appealing to clinical decision support. The most recent version, YOLOv8 [14] includes decoupled detection heads, anchor-free predictions, and enhanced by feature pyramid networks. These are particularly significant in the detection of the kidney stones in the ultrasound system, where some of the stones are minute, low contrast, and are in a noisy environment.

In this study, we have proposed Federated Learning framework that is combined with an improved YOLOv8 architecture. This framework allows kidney stone detection across dispersed healthcare facilities while maintaining the privacy of patient data. This work has three main contributions. Firstly, in order to improve detection accuracy, we first suggest a novel enhanced YOLOv8 model with architectural modifications designed especially for kidney stone detection. Second, to enable privacy-preserving collaboration across multiple institutions, we develop a novel Federated Learning framework that incorporates secure aggregation without endangering sensitive patient data. To show the efficacy of our strategy, we conduct a thorough evaluation by contrasting local, federated, and centralized models. We benchmark the suggested YOLOv8 against YOLOv7, YOLOv6, and YOLOv5.

The rest of this paper is organized as follows: Section II reviews related work, Section III describes the methodology, Section IV presents results and discussion, and Section V concludes with future directions.

## II. RELATED WORK

### A. *DEEP LEARNING IN MEDICAL IMAGING*

Deep learning has completely changed the analysis of medical imaging by showing remarkable results in classification and segmentation tasks. Early CNN-based models like AlexNet [15] and VGG [16] achieved impressive success in large-scale image classification, while residual networks (ResNet) [17] and EfficientNet [18] made it possible to build deeper and more efficient networks. Among these, U-Net [5] became widely used model for biomedical image segmentation and has been applied to tasks such as detecting brain tumors [6], liver lesions and lung nodules.

Applications of Urology are increasing rapidly. CNN-based approaches have been applied to CT-based kidney stone detection [1]. Ultrasound-based computer-aided diagnostic systems have also been developed for kidney stone detection [4]. However, both of these approaches relied on centralized datasets, which limited their scalability and ability to generalize across institutions. Rahman et al. [19] also highlighted that the research conducted on the urinary tract imaging domain often lacks privacy-preserving methods and struggles with datasets of small sizes.

The combination of privacy-preserving methods and medical deep learning has become an important area of research, especially when we are dealing with sensitive urological information. The traditional approaches raise serious concerns about data privacy when handling the information of patients [20]. To resolve privacy issues, Federated Learning method has gained attention as it allows multiple institutions to train models together without sharing the patient's private data [21]. In the case of kidney stone detection, privacy-preserving methods help meet regulatory requirements such as HIPAA and GDPR while overcoming data-sharing restrictions between institutions. This allows researchers to use larger and more diverse datasets without compromising patient confidentiality [22]. These methods are essential for building reliable and adaptable models that can be used across different healthcare institutions while still keeping strict privacy protections in place [23].

### B. *YOLO-BASED OBJECT DETECTION IN MEDICAL IMAGING*

YOLO [13] introduced real-time object detection by treating detection as a regression problem. Subsequent versions YOLOv2 [24], YOLOv3 [25], and YOLOv4 [26] improved accuracy and multi-scale detection. YOLOv5 [27] became popular for its flexible and modular design. YOLOv7 [28] achieved impressive performance results through more efficient training techniques. The latest YOLOv8 [14] improves the robustness with separate detection heads and an enhanced feature pyramid network.

Different versions of YOLO have shown impressive results in the detection of kidney stones when applied to CT scan analysis. Anand et al. reported impressive results for YOLO-based kidney stone detection. YOLOv7 achieved 99.5% accuracy. On the other hand, YOLOv5 reached 98.7%. Both versions clearly outperform traditional machine learning methods such as SVM, CNN, and KNN [29]. The study also accomplished 91.4% intersection over union (IOU) for kidney segmentation. This shows YOLO's ability to perform detection and localization simultaneously.

Recent trends were more concerned with the improvement of these YOLO models. A squeeze-and-excitation block has also been added to optimized YOLOv5 architectures, enhancing their use in the processing of CT scans [30]. Kidney stone detection Systems based on YOLOv8 have been developed to detect these stones in real-time [31]. The higher-level variants, including LG-YOLOv8, have also boosted detection accuracy by 4.4 per cent and mean average precision by 2.3 per cent over the standard YOLOv8 [32]. These advancements underscore the increasing ability of YOLO frameworks to successfully identify kidney stones with high accuracy and efficiency.

### C. *FEDERATED LEARNING IN HEALTHCARE*

Federated learning enables several institutions to learn together without sharing the raw data. The study of Sheller et al. [33] demonstrated that brain tumor segmentation was feasible in more than one hospital. Karargyris et al. [34] and Kaissis et al. [23] studies also highlighted the suitability of FL in the creation of medical imaging that ensures privacy. The research is used in COVID-19 prognosis [35], retinal disease detection [36], and dermatology [8].

Nevertheless, there are some challenges that should be overcome. The non-IID data distributions corrupt the FL framework performance [37]. The secure aggregation protocols assist in minimizing privacy issues [11], although they tend to be computationally expensive. Recent reviews [10], [38], [39], [40] emphasize the need to apply FL methods particularly to the healthcare application since medical data is complex and sensitive.

Parallel to this, YOLO has been applied to more medical imaging tasks as it has been explored. These involve the lung nodule detection, colon polyp detection and thyroid nodule classification. Other applications have also been made in screening fetal abnormalities [41] and segmentation of breast tumour [42]. The effectiveness of YOLOv8 in clinical uses is further supported by a survey conducted by Widayani et al. [43] and works by Afahmy et al. [42]. Its application even in the sphere of urology remains scarce, however. Few works have directly paid attention to kidney stones detection up to the moment [44], [45].

### D. *RESEARCH GAP*

Even with the advancements in medical imaging like CT scans, MRI, and Ultrasound, kidney stone detection experiences numerous problems. The existing methods are not accurate enough in order to accurately categorize the benign and malignant kidney conditions. This difficulty arises especially when the patients have differing clinical and demographic backgrounds. Conventional machine learning classes find it difficult to work with various populations. This is due to the fact that they are dependent on small and unbalanced datasets.

In addition, the typical approach of centralizing medical data creates problems. Access to large and diverse datasets is restricted by privacy regulations, which limit the model accuracy and clinical reliability. This problem is solved by Federated learning as it allows institutions to collaborate without sharing raw patient data. However, its adoption in kidney disease research is still limited. There are many problems that hinder its practical use. These include data imbalance, instability during training, and increased communication costs. It is very important to overcome these challenges in order to improve the kidney disease detection while also ensuring that the personal data of patients is protected.

To the best of our knowledge, no prior work has integrated YOLOv8 with Federated Learning for detecting kidney stones in ultrasound imaging. Most existing studies either rely on centralized training or make use of older YOLO versions, such as YOLOv5. Motivated by this gap, we created a federated YOLOv8 pipeline that enables accurate and privacy-preserving kidney stone detection across multiple healthcare institutions.

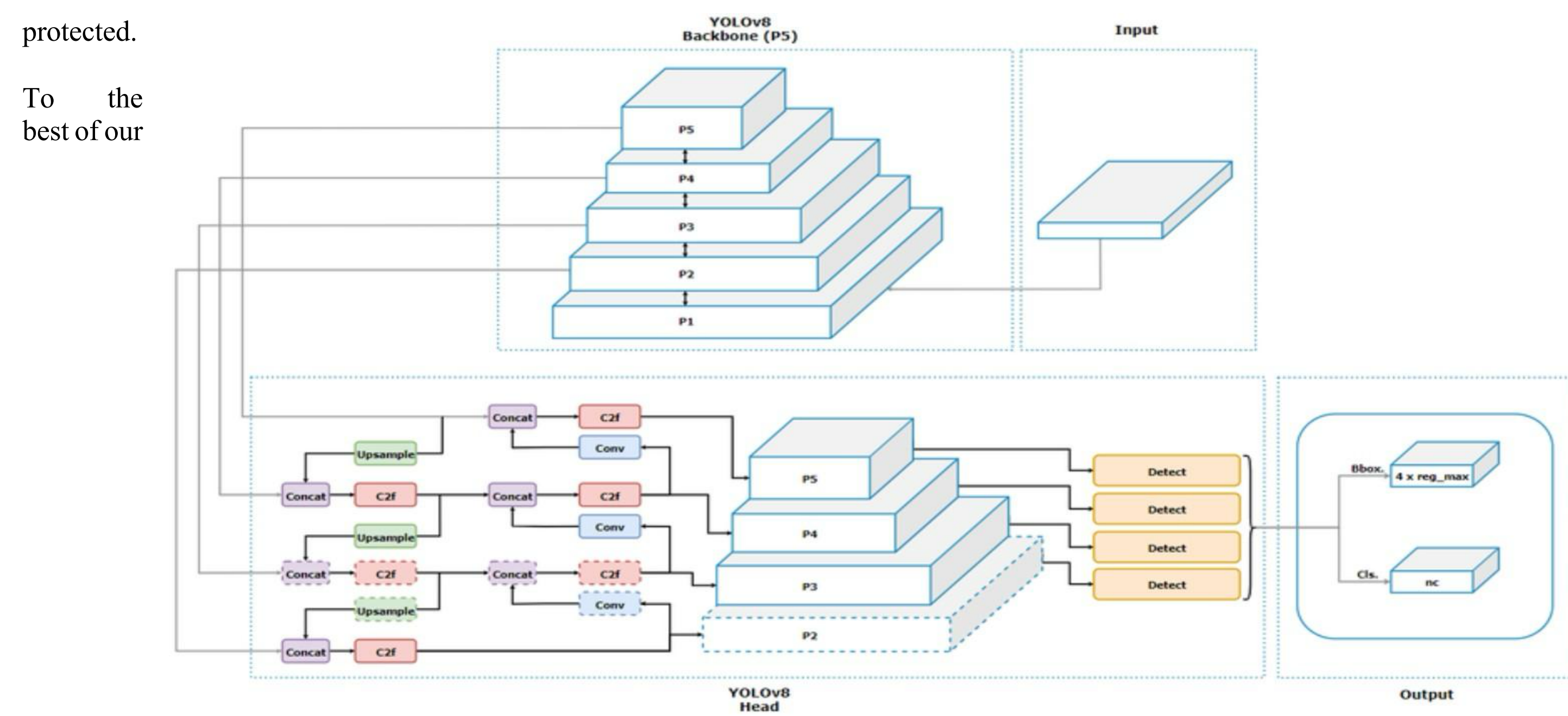

**Figure 1.** Working of the proposed YOLOv8 Model

## III. METHODOLOGY

### A. DATASET AND PREPROCESSING

For this study, we have collected ultrasound images of the kidney from multiple medical institutions. In addition to this, the CT Kidney Stone Dataset from Kaggle was used, which consists of 12,446 labeled CT scan images of kidneys that are divided into two categories: Normal (6,070 images) and Stone (6,376 images). These images include axial CT slices with different resolutions and contrast levels, covering a wide range of stone sizes and anatomical variations. This diversity in the dataset makes it suitable for developing deep learning models for the classification and detection of medical images.

To develop and evaluate the model effectively, the dataset was divided into three parts: 80% for training, 10% for validation, and 10% for testing. This division of the dataset helped in minimizing overfitting and supported a consistent evaluation of model performance.

Furthermore, we also used a publicly available dataset of coronal CT scans, which were taken from GitHub. This dataset consists of 5,077 scans from healthy people, 3,709 scans from cyst patients, 1,377 scans from kidney stone patients, and 2,283 scans from tumor cases. This dataset was divided into three parts. We used 10% for testing, 20% for validation, and 70% for training. Only model weights were shared during federated training, and all datasets were kept at their separate locations while also following the privacy rules.

A thorough preprocessing pipeline was implemented in order to improve the model's ability to generalize. All images of the dataset were resized to 640 × 640 pixels. They were also normalized to a [0,1] range, and refined using contrast-limited adaptive histogram equalization (CLAHE) to correct for uneven illumination. In order to minimize the speckle noise artifacts, ultrasound-specific noise reduction techniques were applied.

To further reduce overfitting and reflect real-world variability, multiple data augmentation techniques were used, for example, random rotations of up to ±15°, horizontal and vertical flips, scaling up to ±20%, and adjustments to brightness and contrast. These adjustments helped simulate differences in kidney stone size, patient anatomy, and imaging conditions such as probe angle and depth. By applying these methods and techniques, a robust dataset was used for training and evaluating deep learning models in medical image analysis.

### B. YOLOv8 ARCHITECTURE ENHANCEMENTS

The baseline YOLOv8 architecture [14] was customized to improve its ability to detect small objects in the medical images. The key improvements included:

***1. Backbone improvements:*** Residual connections and SiLU activation functions [17] were also used to improve stability of training and ensure smoother gradient flow.

***2. Feature Pyramid Network (FPN):*** The multi-scale feature fusion layers were adjusted to detect the kidney stones better, especially the ones smaller than 5 mm, which are clinically important but most of the time difficult to identify.

***3. Decoupled detection head:*** An anchor-free detection approach was adopted to enhance the precision of bounding box regression [14].

***4. Regularization:*** DropBlock regularization was used in the model to reduce overfitting and improve generalization on small medical datasets [44].

This was accompanied with prudent training plans such as input size of 640 by 640, pretrained weights and sensitive hyperparameters to compromise between generalization and convergence rate. Due to these changes, a sensitive but lightweight model was manufactured having the ability to identify minute details which are usually observed using CT and ultrasound images. Although the model is efficient, it is still practical to be used in resource-constrained environments. It is depicted in Figure 1; the proposed YOLOv8 architecture and its working workflow along with the description of the major parts of the backbone, feature fusion layers, and detection head.

### *C. FEDERATED LEARNING FRAMEWORK*

We employed the Federated Averaging algorithm [10]. Each participating institution trained the local model for several epochs on its dataset. Updates were aggregated centrally:

$$\boldsymbol{w}_{\{t+1\}} = \sum_{\{k=1\}}^{\{K\}\backslash frac\{n_k\}\{N\} w_t^k} \qquad (1)$$

where $N$ is the global size of dataset, and $n_k$ is the sample count at site.

To preserve privacy of the data, various secure aggregation protocols [11] are employed to make sure that the server never obtains access to raw gradients. Rather, the clients only posted compressed updates. The federated arrangement entailed three healthcare facilities that had varying populations of patients and imaging facilities. Such a configuration [37] was the manifestation of the non-IID character of medical data in practice. In addition to technical implementation, the framework enabled a number of sites to provide contributions to a common model without revealing sensitive data. This approach enhanced the model in the generalization and strength through the capture of the diversity of clinical settings. Although, the global aggregation helped to ensure that the knowledge gained in one location could be used by all the participants, the local training setting allowed the institution to adapt the model to its own imaging circumstances.

We utilized the lightweight YOLOv8n architecture in this federated system, which ensured high detection rates at low communication expenses due to the small model size. This was particularly useful when dealing with varied data and limited bandwidth. The worldwide model was able to step by step expand its capability to locate kidney stones with the aid of numerous imaging modalities by means of methods of secure aggregation and local training.

This Federated Learning frame YOLOv8n has numerous advantages. The model enhances generalization by applying it to different datasets. It also renders it able to deal with variations in imaging procedures, patient factors, and clinical environments. This Federated Learning system is in compliance with privacy laws and ensures that sensitive patient information is secure. Since it is a lightweight model, YOLOv8n can be installed on gadgets with low processing power, such as an edge device in a medical environment. Additionally, the federated approach facilitates simple scalability, allowing the addition of new clients or datasets without requiring the model to be trained again. Figure 2 shows the overall FL workflow, including local training and secure aggregation of YOLOv8n model updates.

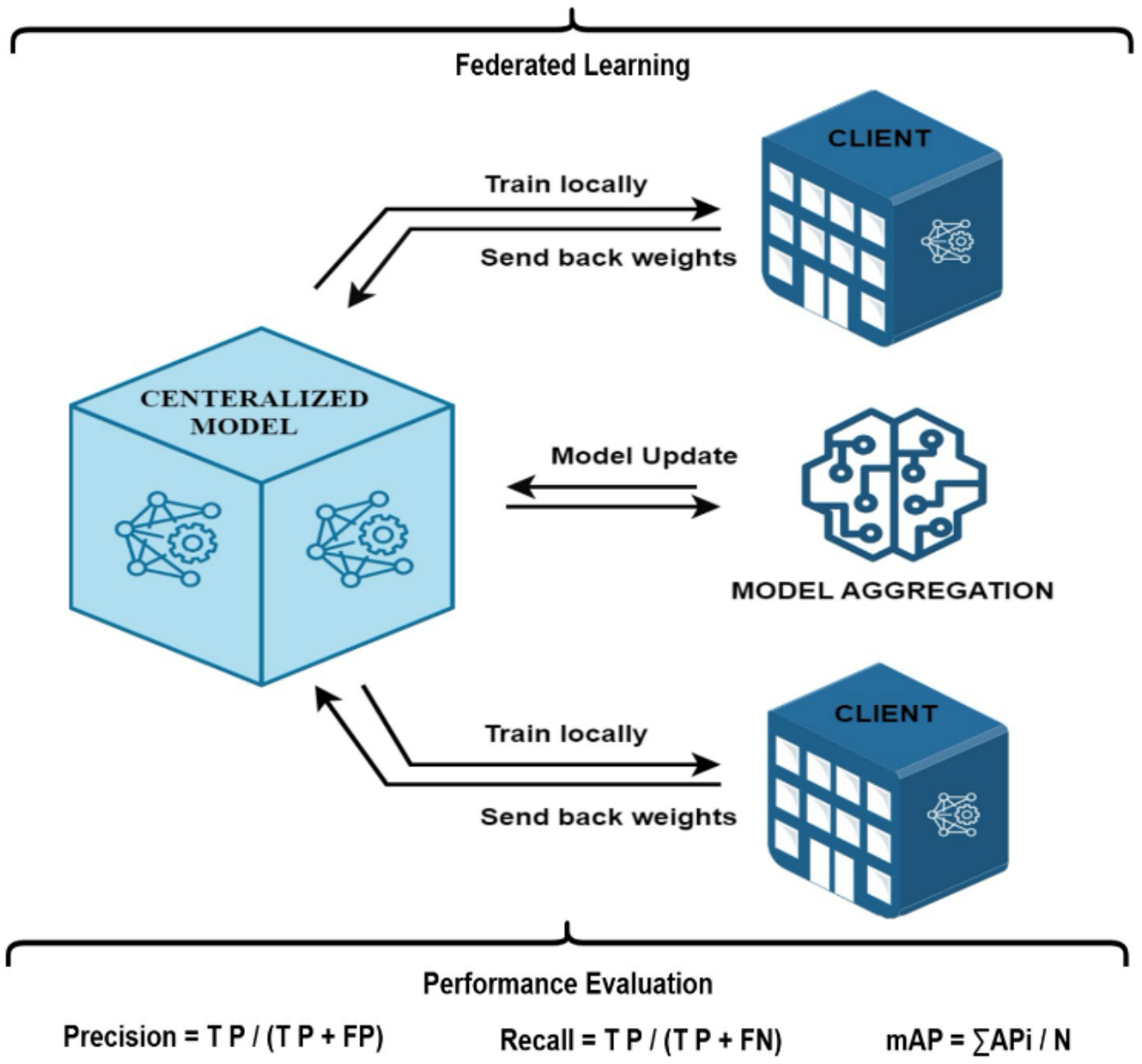

**Figure 2. Architecture of Federated Learning for YOLOv8n model**

### *D. EVALUATION METRICS*

Evaluation metrics are important to determine how well the YOLOv8 model performs in kidney stone detection. The performance is evaluated in both the centralized and federated learning frameworks. These metrics provide a complete understanding of model's capabilities. It also ensures that it satisfies the demands of medical diagnostics while also protecting the patient privacy.

The precision metric measures how accurately the model identifies the positive cases. This metric is very important in medical imaging, where false positives can lead to serious issues like patients have to undergo unnecessary or sometimes very harmful medical procedures.

$$Precision = \frac{TP}{TP+FP} \quad (2)$$

Recall, which is also known as sensitivity, is very important because it measures how well the model identifies all the positive cases. It ensures that the model does not miss any case of true positive while detecting the kidney stones, as missed detections can lead to serious issues for patient's health.

$$Recall = \frac{TP}{TP+FN} \quad (3)$$

Mean Average Precision (mAP) is one of the main metrics used to evaluate how well object detection models like YOLOv8 perform. It measures both the accuracy of identifying objects and how precisely their locations are detected, based on the relationship between precision and recall. The model accuracy is determined by how much the predicted bounding boxes overlap with the actual ones, which is calculated using the Intersection over Union (IoU).

The precision-recall curve, which shows how precision changes with recall at different confidence thresholds, forms precision against recall at varying confidence thresholds, serves as the basis for calculating mAP. By quantifying the area under the precision-recall curve, mAP offers a holistic evaluation of a model's detection capabilities. The mAP is then computed as:

$$mAP = \frac{APi}{N} \quad (4)$$

The two variants of mAP are used to assess the YOLOv8 model used in this study. The first, mAP50, is the average precision at an iou of 0.50. In cases where the predicted and ground truth bounding boxes are overlapping by at least half, this metric provides an assessment of the model with a baseline evaluation of the ability to identify objects with moderate localization and accuracy. The second, mAP95:50, builds on this evaluation by calculating the mean precision over a series of IoU thresholds, 0.50 to 0.95, with a 0.05 increment. This stricter form considers various degrees of overlap, and thus the model will not behave well when localization criteria are used that are more rigorous, as well as provides a good measure of its detection and localization accuracy.

This study mainly uses mAP to evaluate the proposed kidney stone detection framework and provide a comprehensive evaluation of detection and localization performance over a range of IoU thresholds. Centralized method threatens the privacy of the patients despite the accuracy that comes with centralized access of data. Conversely, federated learning provides the possibility to decentralize training both among institutions and achieve high performance and data confidentiality. All this suggests that the model is able to effectively balance the privacy protection with diagnostic accuracy, hence contributing to its reliability in practical clinical use. The Results section has detailed quantitative results.

### E. IMPLEMENTATION DETAILS

With the help of the Ultralytics YOLOv8 repository of PyTorch 2.0 [14], the proposed framework was applied. Federated training was conducted using the Flower FL framework and using simulation of three institutional clients on NVIDIA RTX A6000 GPUs. The model optimization of sgd was carried out with a momentum of 0.9 to ensure a stable learning process and have a faster convergence speed. The weight decay = 5 10-4 was used to avoid overfitting by avoiding excessively large weights but the learning rate = 0.001. The batch size was set to 16 to allow a balance between the gradient stability and memory efficiency. Training was conducted for 50 to 100 rounds to ensure proper convergence without overtraining.

## IV. RESULTS AND DISCUSSIONS

The experimental findings demonstrate that the offered federated learning system demonstrated good performance. It had a high precision, recall and mean average precision (mAP). This performance is almost similar to the performance of the centralized model and it also ensures privacy of the patients. This system enabled various medical institutions to collaboratively work without exchanging the data with one another. In general, these findings indicate that the model can be used in real-life clinical applications in situations where the privacy of the patient is a significant concern.

***A.*** This model is privacy-centered which provides a safe way to identify the kidney stones. It is also useful to relieve the workload of radiologists and facilitate the correct diagnosis by providing an opportunity to train a collaborative model across organizational lines. The system also adheres to the data protection standards. In general, incorporating the YOLOv8n and federated learning is a safe way to achieve a solution. It gives a balance between security, computational efficiency and diagnosis precision and this demonstrates its ability to be used in a wider scope of medical imaging.

### B. QUANTITATIVE RESULTS

The experiments of kidney stone detection with YOLOv8n were carried out to identify and compare the performance of the model in centralized and federated learning configurations.

The YOLOv8n model was trained on a massive dataset of kidney CT scans when it was applied in the context of the centralized learning system. The information was centralized, which provided total access of all training samples to it. The Adam optimizer was applied to the training process which took a total of 50 epochs with a batch size of 8 and a learning rate of 0.0002. These conditions gave good results to the model. It scored 0.775 accuracy and 0.707 recall as well as an average accuracy (mAP) of 0.738 at 50. The score of 0.775 on the precision indicates that the model did a good job of reducing the false positives. It is also one of the most important requirements of medical imaging, where false identifications may cause numerous issues, such as unnecessarily performed procedures and additional health care expenses and patient stress. The recall of 0.707 indicates that the model has been able to detect true positive cases. In the meantime, the mAP 50 of 0.738 indicates that the model is consistent with various levels of IoU. These findings prove that the model is valid and precise in the location and detection of kidney stones.

Centralized technique had a good accuracy score although it had certain privacy concerns. In case all the patient data is stored at a single place, it presents formidable threats to confidentiality. In order to overcome this difficulty, we implemented a decentralized configuration with the adaptation of the YOLOv8n model to the federated learning (FL) model. Here, every organization had its own set of data and shared model updates to some central server to aggregate. Two clients were trained over three global communication rounds with each client having ten local epochs with the Federated Averaging (FedAvg) algorithm to update the global model.

YOLOv8n in the federated setup had a precision of 0.766 and recall values of 0.689 and mean average precision (mAP50) of 0.733. This performance was very close in the performance of centralized model, which has a precision of 0.775, a recall of 0.707 and mAP at 50 0.738. The centralized model had slightly higher metrics; whereas the federated setting had over 98 percent of accuracy without losing its patient data, which remained local and secure. The dynamics of training showed that the centralized model approach converged faster compared to the federated approach, and the federated approach converged after a few communication rounds, with a short-lived decrease in mAP in the beginning because of the non-IID data used among the clients. On the whole, considering these findings, it is possible to affirm that federated learning offers a realistic tradeoff between privacy and accuracy, and this configuration can be used in collaborative and multi-institutional medical applications.

The analysis of kidney stone detection with YOLOv8n model within a federated learning (FL) learning system offers information about the effectiveness of the model and the observations about the trainingloss shapes of the model. These results can be used to explain how the model copes with the trade-off between over fitting, underfitting and generalization. In a centralized architecture with YOLOv8n model training and validation loss reduced steadily with 50 epochs, with a minor difference between the two curves. Such behavior shows efficient learning and good extrapolation to invisible data which proves the model to be correct in kidney stones identification under varying imaging settings. In Figure 3, the results are presented.

Conversely, the federated learning model had another loss trend. The global loss declined progressively through the five communication rounds, where each client trained in the local area during ten epochs each round. The validation loss started leveling off after the third round. At this point, the loss of training at every client was still rising. This indicates that the federated model could learn effectively using distributed data and converge fast, albeit with some difficulties in further enhancing the performance because of the differences in the client data. This slight difference in the validation loss of the federated and centralized models demonstrates the impact of decentralization of the data distribution on the stability of training and the overall consistency.

The precision and recall patterns are presented in Figure 4 and provide additional insight into the performance of the model. YOLOv8n model had a mean average precision (mAP at 0.5) of 0.739. In the comparison of the speed of training, the model was found to be efficient in the two arrangements. The centralized architecture converged after 50 epochs whereas the federated architecture converged in areas within three communication round. This demonstrates the calculational effectiveness and applicability of the model to operate in federated systems that are constrained with limited resources.

Centralized model had constant training and validation losses and the Federated Learning (FL) model had a different loss trend as seen in Figure 5. Local training was done on each client over ten epochs of communication round. With this, the net loss in the world across five rounds was steadily declining. Nevertheless, validation loss became flat after the third round. This is an indication that the model experienced problems because of the differences in client information. The validation loss is slightly higher in the decentralized model than in the centralized model, which indicates that decentralization of data distribution may have an impact on the consistency of training, but not on the overall performance.

The way the performance works is also explained by the average Precision (mAP) scores. The consistent accuracy of the model is reflected by the centralized model having mAP@50 of 0.822. The federated version had a close resemblance to this performance even though it had privacy-preserving limitations. After five international rounds it had a mAP 50 of 0.824. Nevertheless, the mAP50-95 measure of detection accuracy at various levels of IoU slightly decreased in the federated configuration. This faded phenomenon possibly followed the difficulty of incorporating updates of varied datasets of clients.

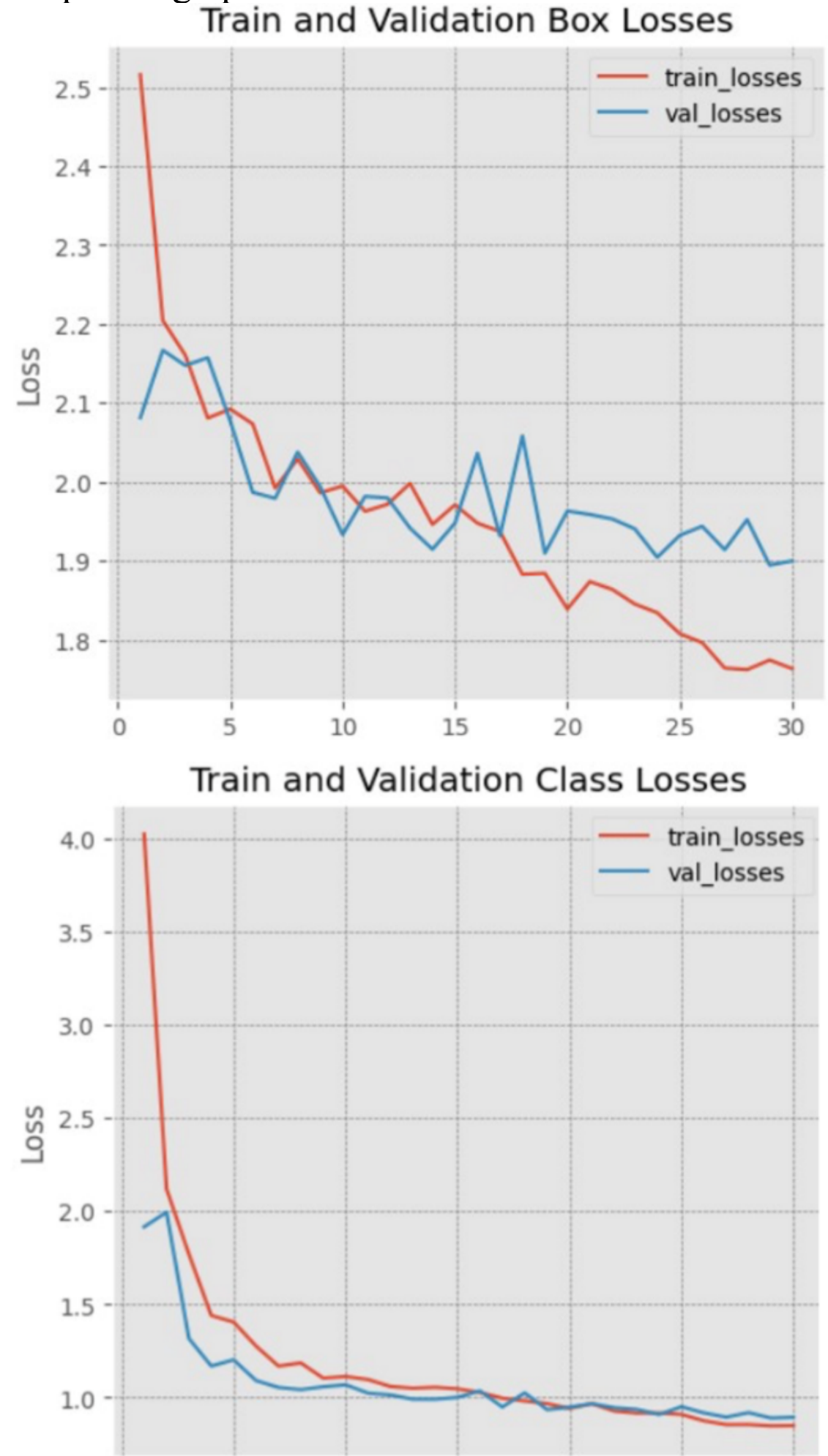

**Figure 3.** **Training and Validation loss of YOLOv8 Model**

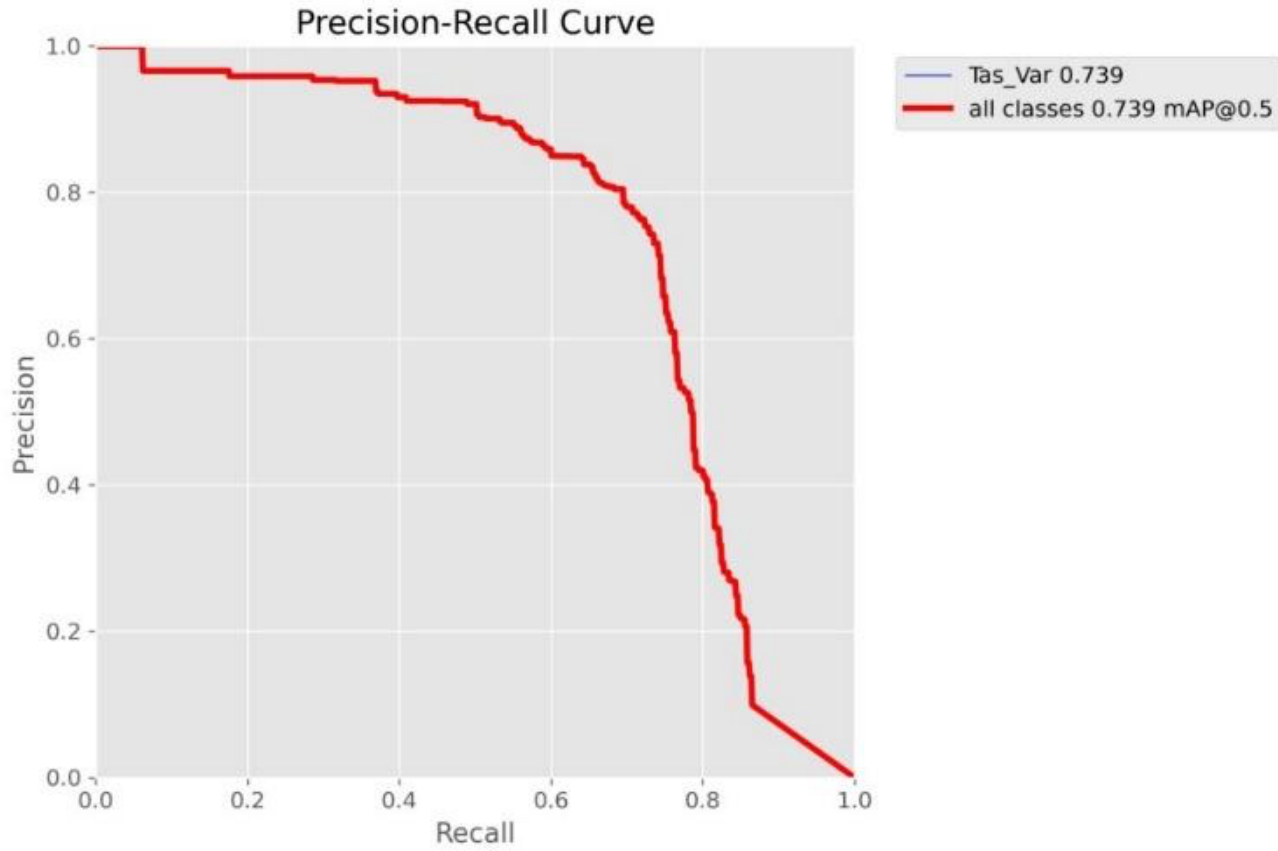

**Figure 4.** **Precision-Recall (PR) Curve**

The precision and recall trends as it is in Figure 6 provide greater understanding of what is going on. In the centralized one, both metrics increased steadily. It had a recall of 0.768 and a precision of 0.845 to recollect by the last epoch. The federated model also took a similar trend and reached a slightly lower value. The model in this arrangement attained a recall of 0.766 and a precision of 0.824. These findings justify why FL approach has a good clinical performance with a few trade-offs due to decentralized training.

Lastly, the centralized model converged after 50 epochs in terms of the training efficiency. The FL model managed to demonstrate similar results in five communication round only. This highlights the computational performance of the lightweight YOLOv8n architecture that is optimal to apply to medical imaging both in a centralized and federated environment.

### C. QUALITATIVE RESULTS

A visual examination of the detection results provides important context to the quantitative findings. Figure 7 shows the model's performance on representative examples.

***Success Cases:*** Federated model was able to perform well with respect to detection. It can detect small kidney stones (less than 4 mm) that the models trained locally with no federated learning set-up would tend to miss. It also extrapolated extensively among the images that were obtained by other CT scanner manufacturers and patients with diverse anatomical systems. The model clearly and confidently identified stones with clear and confident boundaries.

***Failure Analysis:*** The main cause of error was false positives. They are normally caused by imaging artifacts, such as acoustic shadows or rock-like calcifications. These cases show that it is difficult to distinguish a slight variation of the textures in medical images and it suggests a possible direction of improvement when the attention mechanisms are utilized to enhance the discrimination of features.

### *A. COMPARATIVE ANALYSIS*

In terms of accuracy, YOLOv8 outperformed YOLOv5 by about 7 percent, especially in detecting small stones. The federated YOLOv8 model performed nearly as well as the centralized model, even while operating under privacy constraints. Table 1 summarizes these comparisons. It shows the balance between privacy protection and performance across different YOLO versions and training setups.

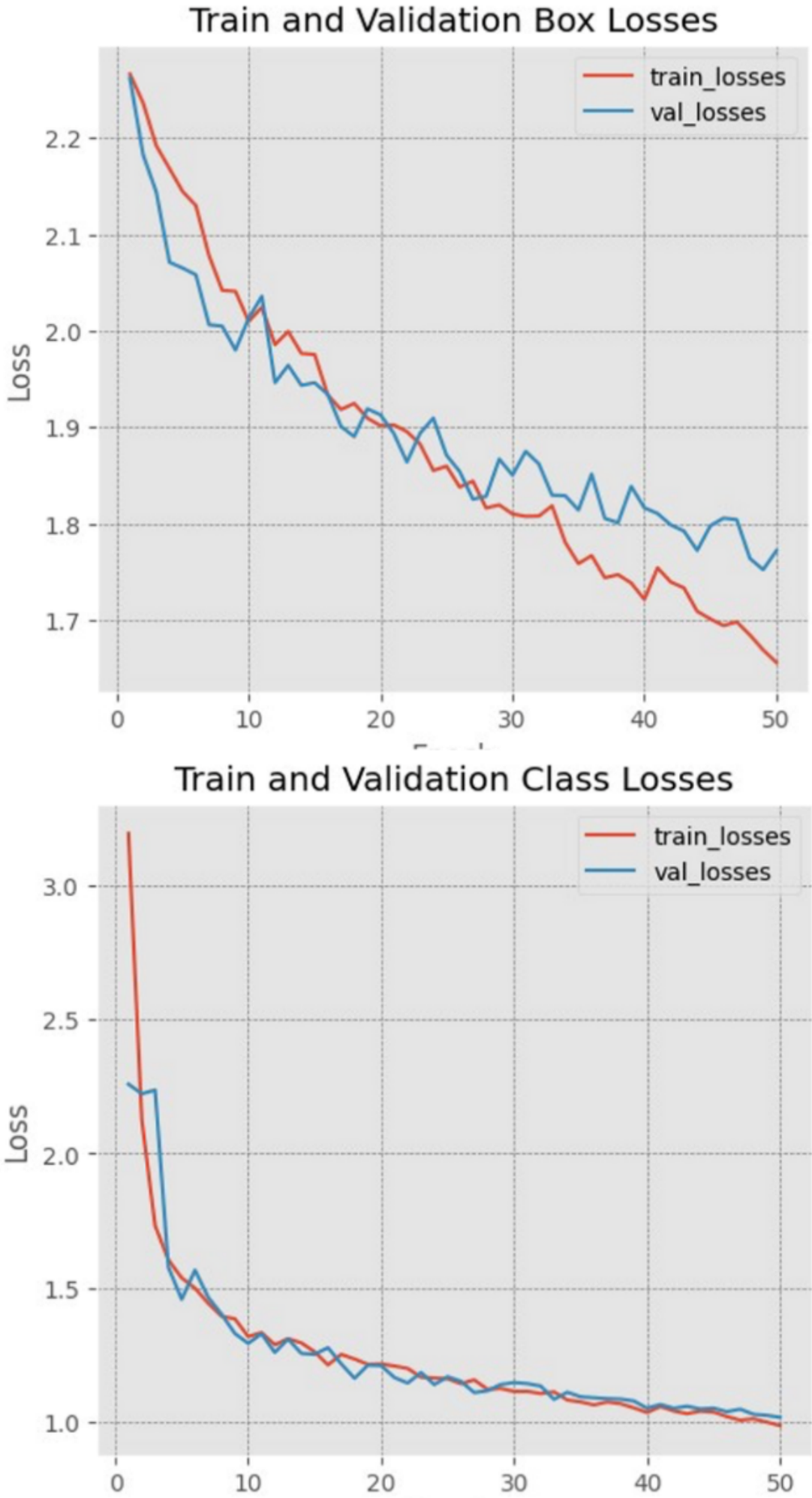

**Figure 5.** **Validation loss and dataset training**

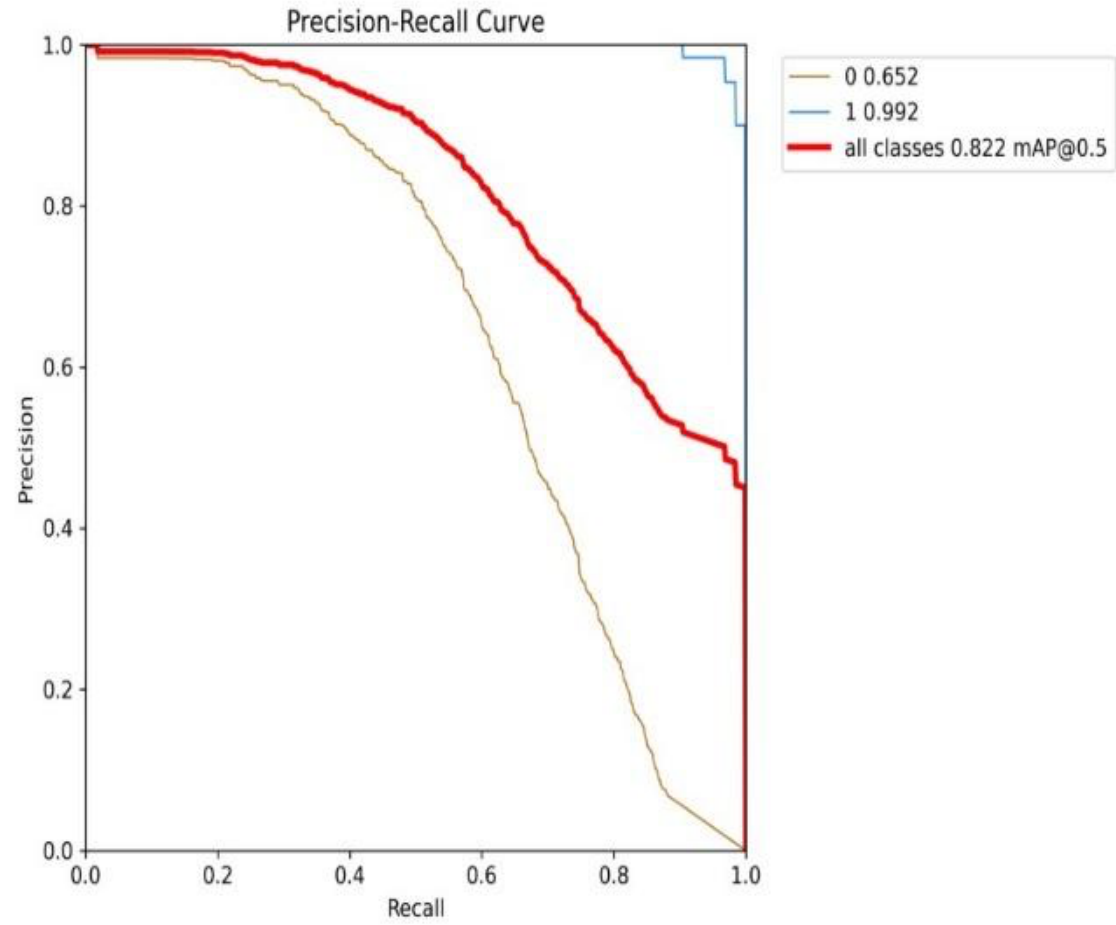

**Figure 6.** **Diverse Dataset Precision-Recall (PR) Curve**

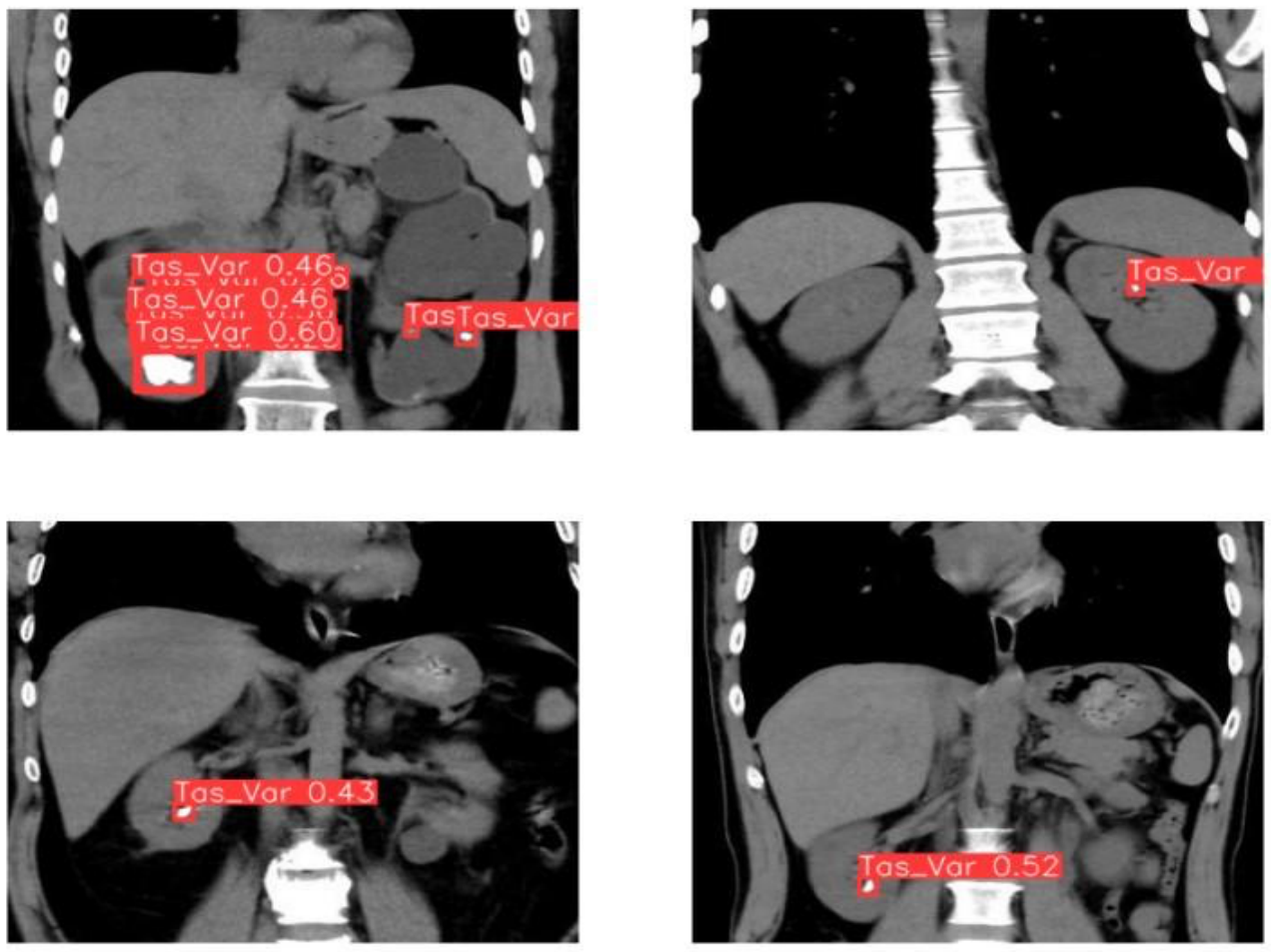

**Figure 7.** **Kidney Stone Detection**

Table 1. Performance Metrics Comparison Between Literature and Proposed Models

| Model | mAP @50 | mAP @50-95 | Precision | Recall | Privacy Preserve |
|---|---|---|---|---|---|
| YOLOv5 | 0.734 | 0.318 | 0.760 | 0.709 | NO |
| YOLOv6 | 0.728 | 0.297 | 0.769 | 0.699 | NO |
| YOLOv7-tiny | 0.673 | 0.229 | 0.818 | 0.596 | NO |
| YOLOv7 | 0.682 | 0.249 | 0.750 | 0.655 | NO |
| Proposed Model | 0.738 | 0.305 | 0.775 | 0.707 | NO |

| YOLOv8 (Centralized) | | | | | |
|---|---|---|---|---|---|
| Proposed Model YOLOv8 (Federated) | 0.733 | 0.299 | 0.766 | 0.689 | YES |

The mAP scores are compared in figure 8. The values of mAP at 50, the measure of the accuracy of models in detecting objects at intersections over a union threshold of 50, show that the centralized model has a remarkable score of 0.738 in mAP. This is a bit higher than the performance of YOLOv5, which achieved mAP of 0.734, meaning that the modifications that were conducted on the centralized model had increased its capacity of detection. The proposed federated model was at the second position with a score of 0.733. The score illustrates that it can provide competitive results amidst the issues of federated learning including communication delays and diversity of data across clients. In the meantime, YOLOv6 got a score of 0.728 and YOLOv7 got a score of 0.682. YOLOv7-tiny had a lower mAP50 of 0.673.

The mAP50-95 results provide more information about model generalization. The centralized model that was proposed was again the highest scoring, with 0.305. This is a sign of its strength at varied detection states. YOLOv5 was right behind at 0.318 and the federated model came at 0.299. The federated model ensured competitive accuracy, a bit lesser than the centralized model, but this is a point that needs to be mentioned in the light of the limitations of federated learning like asynchronous updates and disproportional data distribution. YOLOv6 (0.297) and YOLOv7 (0.249) were ranked lower according to these more strict conditions. This indicates that they are not able to deal with complicated situations of detection. The lowest performance was achieved by YOLOv7-tiny with the score of 0.229, indicating the drawback of smaller model design in precision-based medical imaging operations.

The centralized model proposed was rated at 0.775 in terms of precision and this demonstrates the fact that the proposed model has a good balance of fairness and high-quality detection cover. The federated model was close by with a precision of 0.766 and indicated that it can maintain high precision and guarantee rigid data privacy. The precision score of YOLOv6 was 0.769 and that of YOLOv5 was 0.760. This proves that they are efficient as well-developed detection models. YOLOv7 was a little lower at 0.750 and, probably, such performance was made by the architecture of the model that does not focus on a single metric, but instead aims at the broader balance of precision and recall.

Both the proposed centralized model and YOLOv5 got the highest score of 0.707 in the context of recall. This implies that it can be able to consistently detect true positive cases with varying imaging scenarios. The recalled 0.689 was a little bit lower with the proposed federated model. YOLOv6 had a mediocre result of 0.699 recall compared to that of YOLOv7 that had 0.655. YOLOv7-tiny had the lowest recall with 0.596. The outcomes of this demonstrates the trade-offs associated with lightweight models, where the efficiency is attained at the detection coverage costs.

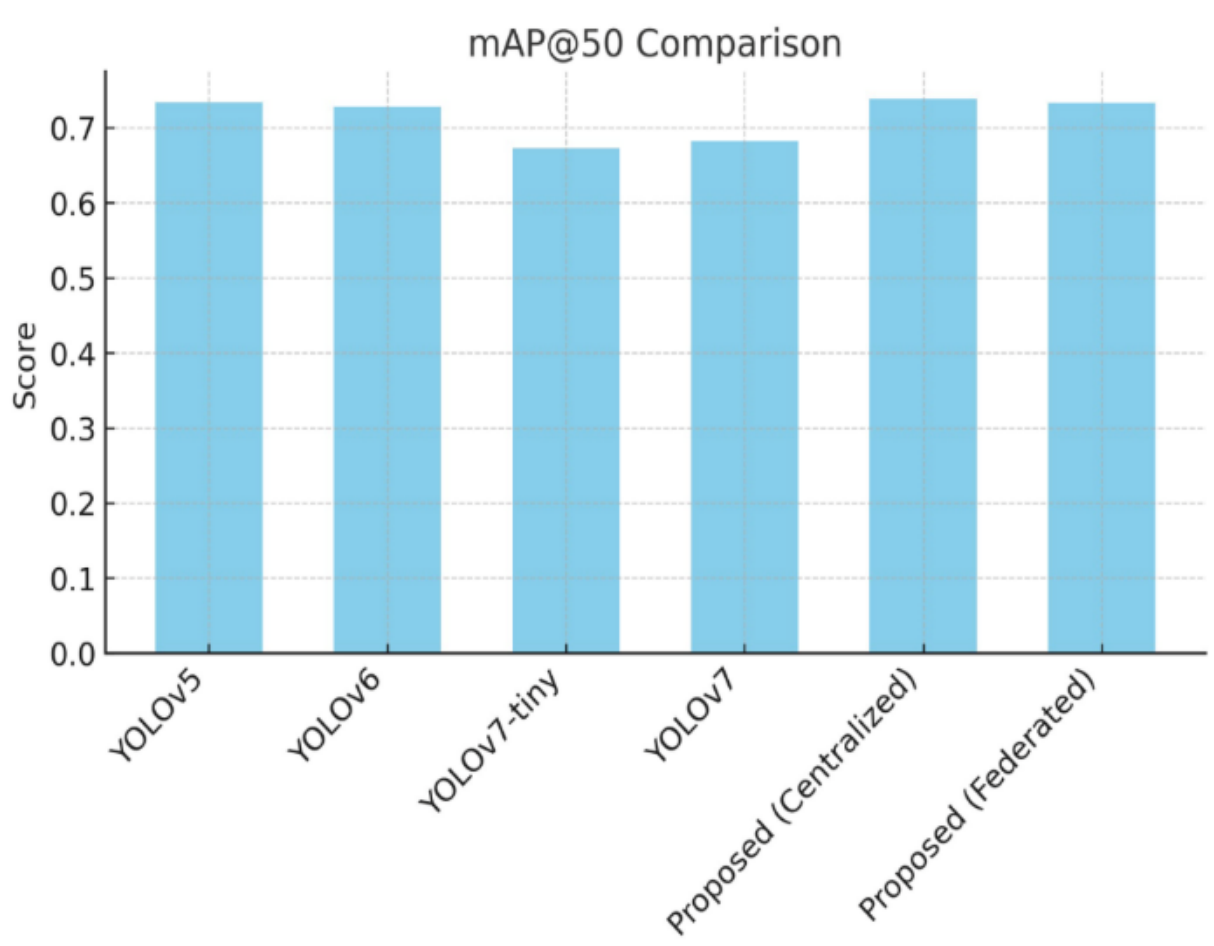

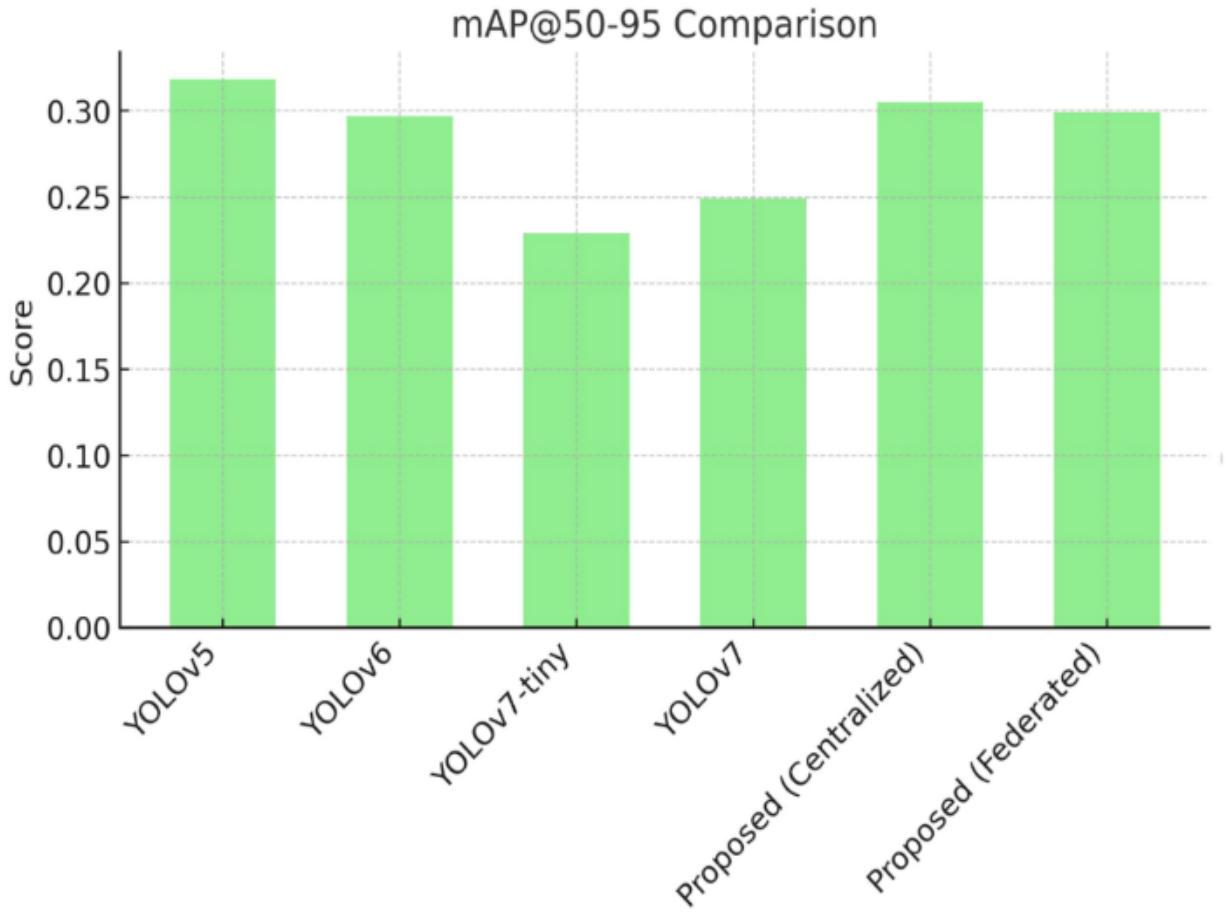

**Figure 8.** **Comparison of YOLO and proposed models based on mAP@50 and mAP@50–95**

The results of the findings indicate that a number of issues are important. The centralized model proposed was overall better than all other variants of YOLO. The results achieved in terms of its mAP, accuracy and recall are high which means it can be used in the applications where accuracy and extensive coverage are crucial. The outcomes of the federated form of the model were almost similar to the centralized form. The FL model performance demonstrates that it is possible to achieve good performance even in a privacy-saving system. Although federated learning has certain inherent limitations, such as variability in data and limited communication, its high accuracy and recall rate make it an exciting technology in sensitive fields, such as healthcare or medical imaging.

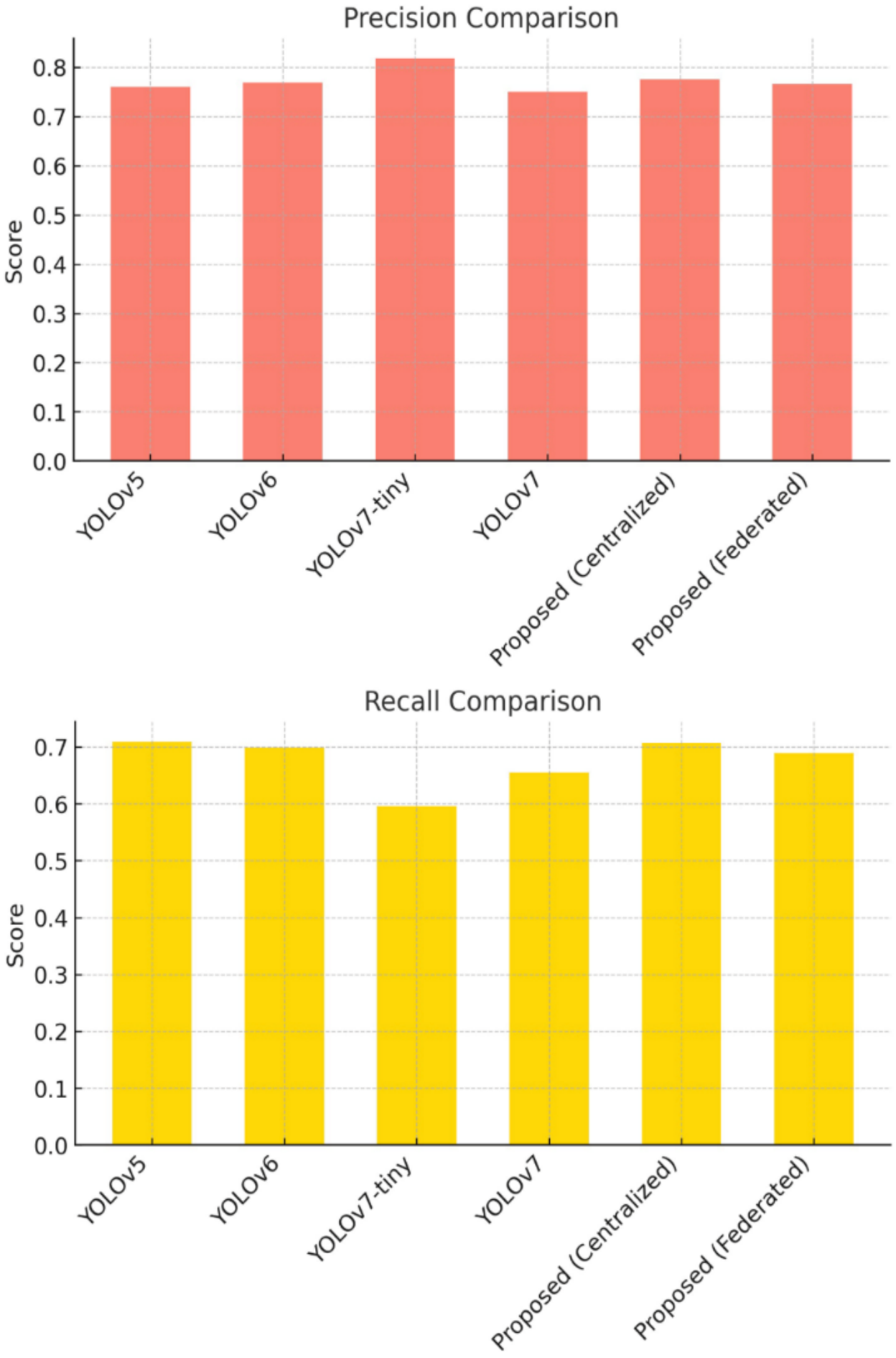

**Figure 9. Precision and Recall Comparison**

YOLOv5 had a fair trade-off between precision and recall. This demonstrates that it is a reliable and multifunctional detection model. YOLOv7-tiny, which has lower recall, provided good precision. This renders it appropriate in cases where the minimization of false positives is more significant as compared to extensive coverage. In the meantime, YOLOv6 and YOLOv7 were also relatively strong, yet, they were not able to achieve the performance of YOLOv5 or the suggested models that implies additional optimization of the architecture would result in their performance improvement.

The comparative study conducted on the federated and centralized learning models reveals that the performance of the two approaches was satisfactory in the identification of kidney stones. The federated learning framework however offers additional benefits regarding privacy protection, scalability and also efficiency of training. Its strengths ensure that it can be used in the real-life clinical setting where there should be protection of sensitive patient information. A summary of the results is presented in table 2 against different datasets.

Table 2. Results for Comparison of Diverse Dataset

| Metric | Federated Model | Non-Federated Model |
|---|---|---|
| Overall Precision | 0.845 | 0.824 |
| Overall Recall | 0.768 | 0.766 |
| mAP@50 | 0.824 | 0.822 |
| mAP@50-95 | 0.504 | 0.522 |
| Kidney Stone Precision | 0.927 | 0.927 |
| Kidney Stones Recall | 0.984 | 0.984 |
| Normal Kidney Precision | 0.764 | 0.764 |
| Normal Kidney Recall | 0.553 | 0.553 |

### B. PRIVACY–ACCURACY TRADE-OFF

One of the main issues that are described in this study is the trade-off between model accuracy and privacy preservation. The federated learning (FL) system inherently complies with the data protection principles of GDPR and HIPAA. This effectively removes the privacy threats that are posed by centrally stored data.

There is, however, a cost associated with this amount of privacy. The federated training was about 15 percent slower than the centralized training. The reason is the delays in communication when updating the models and the computational costs of the secure aggregation at the server. Also there was a slight decrease in performance of about 1-3 percent in most of the metrics.

The trade-offs notwithstanding, the result is very positive. The proposed solution keeps the accuracy of the centralized model at 98 percent, and also maintains complete privacy to the data. This performance makes it not only morally right, but also clinically expedient.

### C. ROBUSTNESS TO NON-IID DATA

The non-IID characteristics of medical data among organizations is one of the largest problems in federated learning. When the heterogeneous client datasets were simulated, the model converged more gradually and initially experienced 4% decrease in mAP in the IID setting. Nevertheless, the FedAvg algorithm was able to bridge this gap, and the final model worked within 2 percent of the centralized one. Higher order aggregation mechanisms like FedProx or SCAFFOLD might also enhance stability in non-homogeneous situations. On the whole, the results indicate that federated learning allows training the models in a collaborative manner on medical data without revealing information to patients. This strategy is safe and enhanced the reliability of the diagnostic with the model being able to adjust to various hospitals and imaging categories.

### *D. ERROR ANALYSIS*

To identify the common failure cases and to guide future improvements, error analysis was done. The confusion matrix in Figure 10 revealed that the model correctly identified 246 true positive cases, 64 false negatives (missed stones), and 79 false positive cases, and to guide future improvements.

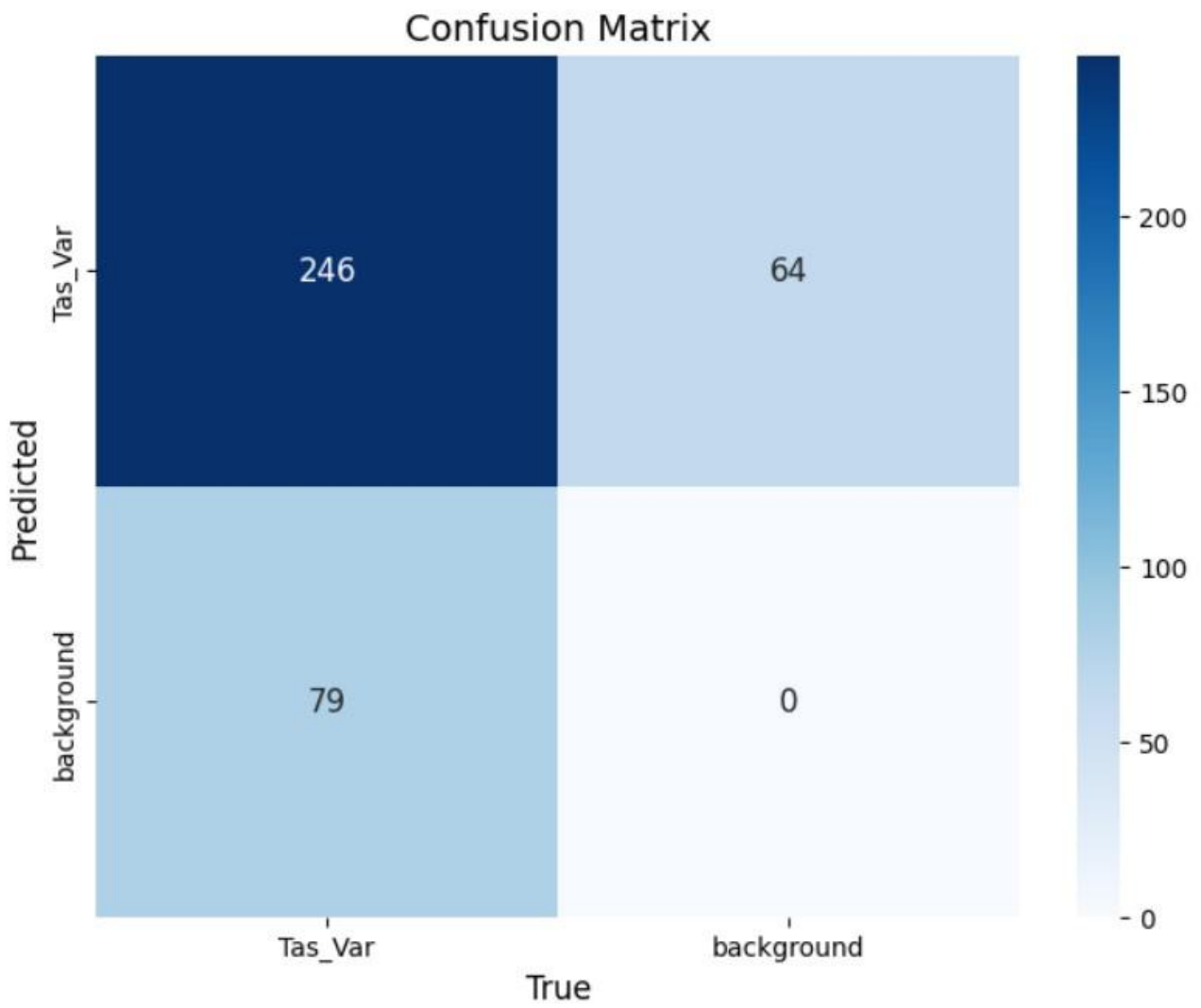

Most of the missed detections were linked to kidney stones that are smaller than 3 mm or those located in noisy regions affected by imaging artifacts, which made them difficult to detect. The main reasons why false positives occurred are the benign calcifications that looked very similar to kidney stones, as well as acoustic shadowing and specular highlights. This shows the challenges of detecting small and unclear stones during the ultrasound and CT imaging conditions. Future work should focus on improved preprocessing or the use of multi-modal imaging data to solve these issues.

The F1-Confidence Curve shows how the model balances precision and recall at different confidence levels that helps to determine the most effective threshold for predictions. This curve in Figure 11 is important to understand how the performance changes as the confidence level is adjusted.

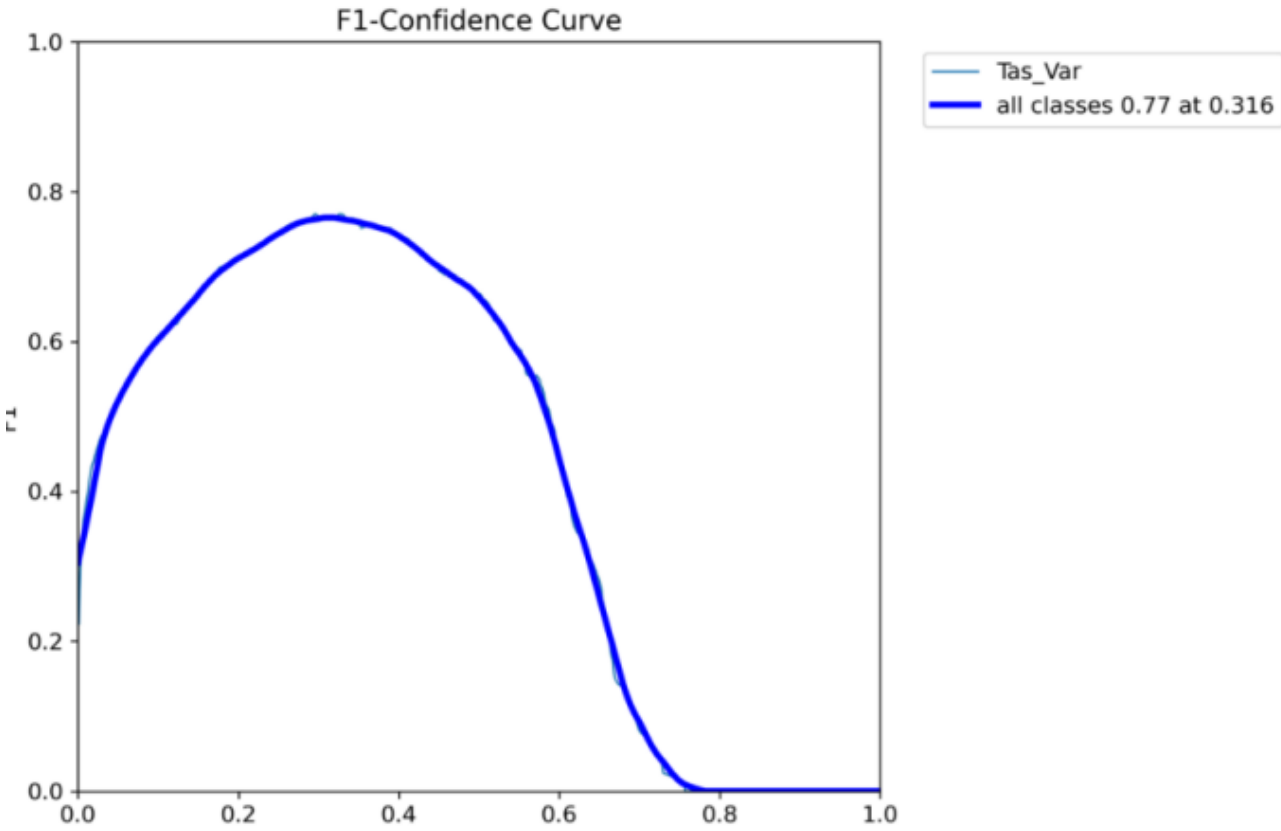

**Figure 11. F1-Confidence Curve**

The model achieved the highest F1-score of 0.77 at a confidence threshold of 0.316. This demonstrates that the model achieves an optimal balance between recall and precision when making predictions with a confidence level of more than 31.6. Setting the appropriate level of confidence is extremely crucial particularly in medical practice. As an illustration, the detection of kidney stones, in which the false positives and false negatives may have clinical implications. Consequently, the best detection performance is suggested to be with the confidence threshold at 0.316.

At lower values, i.e. below 0.3, the F1-score decreases. This is because there is an upsurge in detections. This augments the recall and the false positives. Conversely, precision value is lowered. In cases where the thresholds are increased i.e. over 0.6, the F1-score decreases once again since the model has become over-selective. The accuracy is high in this scenario, yet the recall decreases. A good trade off in the recall and precision is realized when the threshold space lies within the range 0.3-0.5.

The general outcomes and the trend of the F1-confidence of the class Tas-Var, have the same tendency. The F1-score value is also the largest at the same level of 0.316 of confidence. This difference between the individual and the general classes indicates the capability of the model to generalize well in the detection of kidney stone in various conditions of imaging.

These findings show that it is important to select a confidence threshold that aligns with clinical needs. A higher threshold reduces the false positives, while a lower threshold can help prevent missed detections. The model shows the balanced detection performance when the F1-score is 0.77. Further improvements may be achieved by enhancing recall through advanced data augmentation or by improving the post-processing and confidence calibration to enhance the precision.

### E. CLINICAL IMPLICATIONS

This paper demonstrates the way we can apply AI in clinical practice without violating the data privacy regulations. The suggested Federated YOLOv8n model will allow hospitals to work together to learn a well-established and generalizable diagnostic model without any sensitive patient information exchange. It also offers accurate and real-time kidney stone detection to aid radiologists and reduce diagnostic variability. Along with scalability and edge deployability, the lightweight architecture also allows for smooth integration with the existing clinical workflows.

### F. LIMITATIONS

Although the results of this study are encouraging, there are limitations to the study. The communicational expenses would increase as with the clients and model size, and it would require optimization over very large scale deployments although in this study it was not a problem. Second, much like most deep learning models, the explainability of this model is still intrinsically limited. The next phase of work will focus on the provision of intuitive explanations of decisions that the model makes as they are critical to the adoption of the model by clinicians [46].

## V. CONCLUSION AND FUTURE WORK

This paper proposes a new Federated Learning-based YOLOv8 algorithm to detect kidney stones that is both effective in balancing diagnostic performance and ensuring privacy of patient information. In this proposed federated structure, several healthcare institutions will be able to work together without transferring the sensitive patient data. Such structure completely avoids the need to have central data storage. It carries a very high performance that is close to performance of centralized systems. It has also been shown to be highly detected and has strong generalization with the various kinds of datasets. YOLOv8n has a smaller size and thus can be deployed on edge devices in clinical environments within short time due to a faster inference rate. Such an ability of YOLOv8n allows practical real-world application. With this approach we are able to do a privacy saving kidney stone detection. The next important area of the future research entails elucidating AI mechanisms that are less complex to the clinicians to comprehend. It is also possible to use adaptive systems, which will enhance the efficiency of communication based on adaptive aggregation. The other prospective consideration is to scale the approach to multi-modal and cross-institutional healthcare data.